\documentclass[sigconf]{acmart}

\AtBeginDocument{%
  }

\copyrightyear{2024}
\acmYear{2024}
\setcopyright{rightsretained}
\acmConference[RecSys '24]{18th ACM Conference on Recommender Systems}{October 14--18, 2024}{Bari, Italy}
\acmBooktitle{18th ACM Conference on Recommender Systems (RecSys '24), October 14--18, 2024, Bari, Italy}\acmDOI{10.1145/3640457.3691700}
\acmISBN{979-8-4007-0505-2/24/10}

\begin{document}

\title{\emph{Stalactite}: Toolbox for Fast Prototyping of Vertical Federated Learning Systems}

\author{Anastasiia Zakharova}
\email{nastyazakharova.nz@gmail.com}
\orcid{0000-0002-7624-6790}
\affiliation{
  \institution{ITMO University}
  \city{Saint-Petersburg}
  \country{Russian Federation}
}
\author{Dmitriy Alexandrov}
\email{mr.alexdmitriy@mail.ru}
\orcid{0000-0002-3494-5315}
\affiliation{
  \institution{ITMO University}
  \city{Saint-Petersburg}
  \country{Russian Federation}
}
\author{Maria Khodorchenko}
\email{mariyaxod@yandex.ru}
\orcid{0000-0001-5446-5311}
\affiliation{
  \institution{ITMO University}
  \city{Saint-Petersburg}
  \country{Russian Federation}
}
\author{Nikolay Butakov}
\email{alipoov.nb@gmail.com}
\orcid{0000-0002-2705-1313}
\affiliation{
  \institution{ITMO University}
  \city{Saint-Petersburg}
  \country{Russian Federation}
}
\author{Alexey Vasilev}
\email{alexxl.vasilev@yandex.ru}
\orcid{0009-0007-1415-2004}
\affiliation{
  \institution{Sber AI Lab}
  \city{Moscow}
  \country{Russian Federation}
}
\author{Maxim Savchenko}
\email{savvvan@gmail.com}
\orcid{0009-0003-4180-9869}
\affiliation{
  \institution{Sber AI Lab}
  \city{Moscow}
  \country{Russian Federation}
}
\author{Alexander Grigorievskiy}
\email{alex.grigorievskiy@gmail.com}
\orcid{0000-0003-4815-0641}
\affiliation{
  \institution{Independent Researcher}
  \city{Helsinki}
  \country{Finland}
}

 \renewcommand{\shortauthors}{Anastasiia Zakharova et al.}

\begin{abstract}
Machine learning (ML) models trained on datasets owned
by different organizations and physically located in remote databases
offer benefits in many real-world use cases. State regulations
or business requirements often prevent data transfer to a central location, making it difficult to utilize standard machine learning algorithms. 
Federated Learning (FL) is a technique that enables models to learn
from distributed datasets without revealing the original
data. Vertical Federated learning (VFL) is a type of FL
where data samples are divided by features across several data
owners. For instance, in a recommendation task, a user can interact with various sets of items, and the logs of these interactions
are stored by different organizations. In this demo paper, we present
\emph{Stalactite} - an open-source framework for VFL that provides the
necessary functionality for building prototypes of VFL systems. It has
several advantages over the existing frameworks. In particular, it allows
researchers to focus on the algorithmic side rather than engineering and to easily deploy
learning in a distributed environment. It implements several VFL algorithms and has
a built-in homomorphic encryption layer. We demonstrate its use on a real-world recommendation datasets.
\end{abstract}



\maketitle

\section{Introduction and motivation}
In the last decade, many large organizations have positioned themselves as ecosystems, i.e., groups of companies that cover all user needs. Therefore, one of the possible options for improving the quality of recommendations is to enrich models with information from a related company with the same users. Often, companies from the same group may have different owners, so direct original data exchange may not be possible due to legal aspects. \emph{Federated Learning} (FL) is usually used to exchange information and enrich models.

The term \emph{Federated Learning} (FL) was coined in~\cite{mcmahan_17a} to describe a setup where different data owners contribute distinct data samples to an overall system. This type of FL is called horizontal FL (HFL). In scientific literature, the term \emph{Federated Learning} typically refers to HFL. In contrast, Vertical Federated Learning (VFL)~\cite{liu_24,khan_2022} is a setup where data is divided by features. Recognition of VFL is gradually increasing due to its relevant practical use cases. For example, in recommender systems~\cite{cui_2021}, different platforms may collect various parts of user interaction data. A closely related concept is Cross-Domain Recommender Systems~\cite{chen_22, samra_24}. VFL has applications in finance~\cite{luo_23,chen_2021}, healthcare~\cite{shan_23}, advertising~\cite{li_2023}, etc. Split learning~\cite{poirot_2019} is also a type of VFL. In this work, we have focused on the vertical federated learning case.

The development of FL software toolboxes has historically focused on horizontal FL, often leaving the needs of researchers working on vertical federated learning unmet. Some existing toolboxes offer limited or no support for VFL~\cite{ibm} (IBM). Even when toolboxes support VFL, the support is often limited in scope and requires substantial effort to implement new VFL algorithms. The root cause is that their architecture is primarily built with horizontal FL use cases in mind~\cite{fedml, flower}. 

Additionally, several toolboxes are designed for practical industrial use~\cite{nvidia_flare} (NVIDIA),~\cite{openfl} (Intel), ~\cite{fate,paddlefl} (Baidu), making them challenging for researchers to adopt. These industrial toolboxes are optimized for performance, often at the expense of code readability. Furthermore, they may require industrial-level infrastructure and significant engineering efforts to deploy effectively. In these toolboxes, the convenience of modification and implementation of new algorithms is often sacrificed in favor of speed and security. This trade-off can make it difficult for researchers to adapt these tools for experimental purposes or to integrate novel algorithms easily.

Recently, a specialized VFL toolbox for research, \emph{VFLARE}~\cite{vflare}, has been introduced. It implements numerous VFL algorithms and attacks and contains several VFL datasets. Its current functionality is mainly developed for emulating VFL on a single machine, limiting its usefulness for analyzing algorithms in real distributed deployments. The distributed version of \emph{VFLARE} is still under development. In response to these limitations, we have developed \emph{Stalactite}, a toolbox for fast prototyping VFL systems. The main goal of Stalactite is to allow researchers to focus on algorithms rather than engineering while facilitating the deployment of VFL algorithms in real distributed environments across the internet.

VFL training consists of two phases: data matching and model training. The first phase aims to identify common samples across all participants. Once these common data samples are identified, the second phase involves training the ML model. In VFL, participants are typically divided into server and client roles. The server party usually holds the labels and controls the training process. Client parties contribute their data to the training process. Unlike HFL, where model parameters or gradients are exchanged between participants, VFL involves exchanging representations of distributed features or predictions. Stochastic Gradient Descent (SGD) or its variants are commonly used as optimization algorithms. Various techniques are employed to ensure data privacy among participants. Consequently, a single training iteration may require multiple exchanges between the server and clients~\cite{logreg}. Given these complexities, a VFL toolbox must provide a flexible way to modify the main training loop to satisfy the specific requirements of different privacy-preserving techniques.
 
The main features of Stalactite are as follows:
\begin{enumerate}
\item Well-designed abstractions that separate mathematical concepts from message exchange logic, facilitating the easy translation of VFL algorithms into code.
\item Multiple execution modes: multi-thread, multi-process, and distributed, with seamless switching between modes without requiring code modifications.
\item Convenient debugging of algorithm implementations, enabled by the flexibility in execution modes.
\item Comprehensive logging of payload, exchange time, and machine learning metrics during distributed execution.
\end{enumerate}
\textit{Stalactite} source code is available on GitHub\footnote{\url{https://github.com/sb-ai-lab/Stalactite}}. 
An additional contribution to the paper is the presentation of a new real-world open-source dataset SBOL\footnote{\url{https://www.kaggle.com/datasets/alexxl/sbol-dataset}}, which has intersections in users with the dataset MegaMarket \cite{bench}. The characteristics of the SBOL dataset are shown in Table \ref{tab:stats}.

\begin{table}[!htbp]
\caption{\textbf{Statistics of the SBOL dataset for a period of 4 months. The dataset contains information about offers and purchases of banking products by users on certain days.}}
\label{tab:stats}
    \centering
    \begin{tabular}{lr}
   \hline 
        \textbf{Statistics} & \textbf{Total}
        \\ \hline 
        Users & 190 439\\
        Items & 19\\
        Interactions & 1 056 889\\
        Other features & 1 345\\
        \hline

    \end{tabular}
\end{table}

In the current demonstration, the SBOL dataset serves as the primary data source, while MegaMarket contains a subset of users from the main dataset with additional features. We focus on recommending a small set of banking products to users. This example is only a demonstration of Stalactite's capabilities such as exchanging (possibly encrypted) intermediate computations between parties. The framework can be applied to a broader range of recommendation tasks within the VFL formulation. It is capable of handling a larger number of items, for instance, by implementing the recent algorithm described in~\cite{samra_24}. This flexibility allows for more complex and comprehensive recommendation systems to be built using our tool.

\section{Stalactite}

\begin{figure}[h!]
  \centering
  \includegraphics[width=0.9\linewidth]{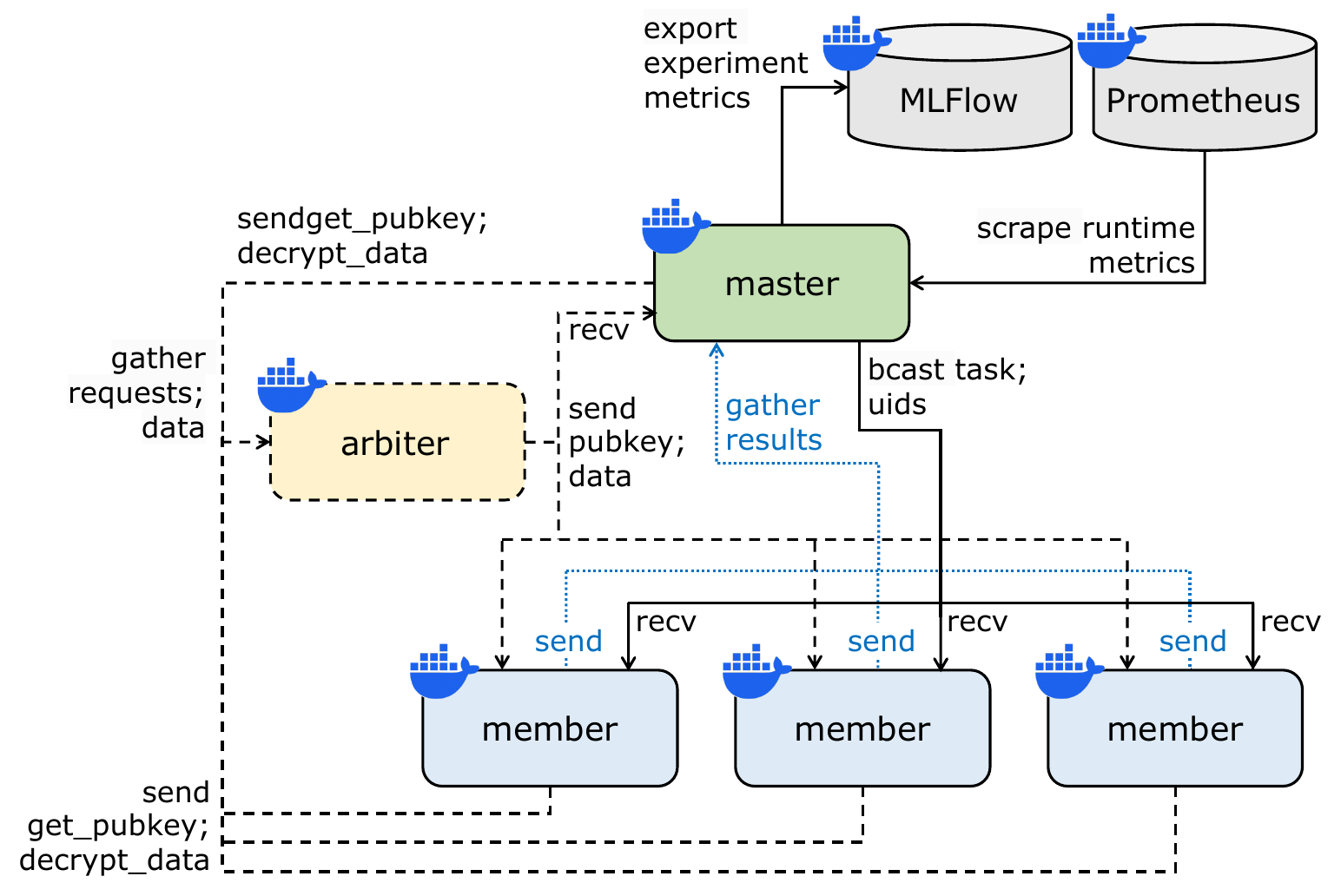}
  \caption{Stalactite architecture with main components and communications between them.}
  \label{fig:arch}
  
\end{figure}

The architecture of the framework with its main components is presented on fig~\ref{fig:arch}.

The master component maintains its part of the data and target labels. It is responsible for matching the records' IDs to form the shared space of rows, synchronizing all iterations in the training process, and calculating the loss. Member component, on the other hand, only holds its dataset and computes forward, and backward passes on its data. The special component Arbiter performs the distribution of encryption keys and calculation of the gradients concerning the master and members. It should be noted that the presence of this component is protocol-dependent and it may be absent if the protocol assumes direct communications between agents. Additionally, there are MlFLow and Prometheus components which are responsible for the collection of the training and inference metrics and statistics, allowing monitoring of the framework's performance and algorithms quality.       

The framework's architecture can be divided into several main layers: communication layer, protocol layer, and models layer. They are implemented in isolation to make the customization and alternations possible if necessary.

\textit{The communication layer} is responsible for organizing data transfers between all participating entities including PartyMaster, PartyMember, and Arbiter. The main entity on this level is the PartyCommunicator which is responsible for a specific implementation of agent's communication while providing a simple MPI-like send/receive interface to the agents.  Currently, the framework offers two different implementations: gRPC-based server-client communication for the distributed setting with Protobuf interfaces and Safetensors serialization; and local in-process thread-based implementation for easy-to-use and easy-to-debug use in IDE. The latter one employs an in-memory queue for sending and receiving data. The combination of gRPC, Protobuf, and SafeTensors has been chosen due to several reasons. First, it saves the volume of data tensors being moved across the network. Second, it is efficient communication over the Internet network (we assume that data silos will be more likely allocated on independent stack holders hardware, not in the same local network). Third, it is flexible in terms of implementing various patterns of communication, including cases where one of the agents can be lost in comparison to traditional communicating frameworks like MPI or gloo optimized for local networks with high-speed connections. At the same time, the local mode strips out all complications caused by distributed settings and makes it easy to concentrate efforts on high-level details of protocol or ML model development and debugging. This architecture enables fast prototyping while preserving seamless switching to a distributed mode when necessary.

\textit{The protocol layer} is responsible for defining the logic of interactions and synchronization between agents, encrypting, and ensuring datasets' non-disclosure for all participants. On this layer, we implement base classes for interactions of the agents in the case of classical ML algorithms, such as linear and logistic regressions, as well as neural networks-based algorithms enabled with a split-learning approach. Homomorphic encryption is also implemented on this layer.

\textit{The models' layer} is responsible for integrating ML models into the framework, regardless of a specific protocol on the previous layer. This layer provides the necessary interfaces for models to be integrated and used by protocol implementations.   

\section{Setup}
Stalactite \footnote{https://github.com/sb-ai-lab/Stalactite} is available open-source on GitHub. Here you will find installation instructions using poetry. The documentation provides detailed information on each of the Stalactite modules. If you have any issues or questions about using the framework, you can check out the source code and contribution guidelines on GitHub.

\section{Demo}
The Stalactite demo illustrates the extensive capabilities of the Stalactite framework through practical applications using independent real-world datasets, SBOL and MegaMarket, which share an ID space in the e-commerce domain. Data from these sources is utilized in a vertical federated learning (VFL) environment to demonstrate how distributed machine learning models can be trained without compromising data privacy. It is achieved by leveraging three cloud-based virtual machines to host distinct network agents. The demo guides through the setup of the repository, configuration of arbitered and arbiterless federated experiments, and execution using the Stalactite CLI to provide an example of machine learning lifecycle management. This includes data synchronization, model training, and result monitoring through integrated tools like MLflow and Grafana. It also provides various advanced features, such as plugin deployment for new algorithm integration and an IDE-supported local debugging mode, positioning Stalactite as a robust solution for VFL algorithm prototyping and distributed applications across diverse computing environments.

\section{Conclusion}
Stalactite provides the opportunity for fast VFL algorithm prototyping and probing. Users can implement ML algorithms, and models and customize the communication protocols for the data exchange between agents with optionally enabled Homomorphic Encryption. The framework allows changing the communication layer with ease(e.g. replacing the gRPC with in-memory exchanges, MPI, etc.) and enables user-friendly debugging for the developing solutions via IDE with easy transfer of those to the deployment via CLI. The proposed architecture of Stalactite allows upgrading it into a fully-fledged industrial VFL framework.

\begin{acks}
 This work was supported by the Analytical Center for the Government of the Russian Federation (IGK  000000D730324P540002), agreement No. 70-2021-00141
\end{acks}
\bibliographystyle{ACM-Reference-Format}
\bibliography{stalactite}

\end{document}